\documentclass[11pt]{article}
\usepackage{times}
\usepackage{url}
\usepackage{latexsym}
\usepackage{amssymb}
\usepackage{afterpage}
\usepackage{naaclhlt2015}
\usepackage{graphicx}
\usepackage{color}
\usepackage{enumitem}

\title{SemEval-2015 Task 10: Sentiment Analysis in Twitter}

\author{
{\bf Sara Rosenthal }\\{\small Columbia University}\\{\scriptsize {\tt sara@cs.columbia.edu}}\\\\
{\bf Saif M Mohammad}\\{\small National Research Council Canada}\\{\scriptsize {\tt saif.mohammad@nrc-cnrc.gc.ca}}\\
\And
{\bf Preslav Nakov}\\{\small Qatar Computing Research Institute}\\{\scriptsize {\tt pnakov@qf.org.qa}}\\\\
{\bf Alan Ritter}\\{\small The Ohio State University}\\{\scriptsize {\tt aritter@cs.washington.edu}}\\
\And
{\bf Svetlana Kiritchenko}\\{\small National Research Council Canada}\\{\scriptsize {\tt Svetlana.Kiritchenko@nrc-cnrc.gc.ca}}\\\\
{\bf Veselin Stoyanov}\\{\small Facebook}\\{\scriptsize {\tt vesko.st@gmail.com}}\\
}

\date{}

\begin{document}
\maketitle
\begin{abstract}
In this paper, we describe the 2015 iteration of the SemEval shared task on Sentiment Analysis in Twitter.  This was the most popular sentiment analysis shared task to date with more than 40 teams participating in each of the last three years. This year's shared task competition consisted of five sentiment prediction subtasks. Two were reruns from previous years: (A) sentiment expressed by a phrase in the context of a tweet, and (B) overall sentiment of a tweet. We further included three new subtasks asking to predict (C) the sentiment towards a topic in a single tweet, (D) the overall sentiment towards a topic in a set of tweets, and (E) the degree of prior polarity of a phrase.
\end{abstract}

\section{Introduction}

Social media such as Weblogs, microblogs, and discussion forums are used daily to express personal thoughts, which allows researchers to gain valuable insight into the opinions of a very large number of individuals, i.e., at a scale that was simply not possible a few years ago. As a result, nowadays, sentiment analysis is commonly used to study the public opinion towards persons, objects, and events. In particular, opinion mining and opinion detection are applied to product reviews~\cite{Hu04}, for agreement detection~\cite{conf/naacl/HillardOS03}, and even for sarcasm identification~\cite{conf/acl/Gonzalez-IbanezMW11,liebrecht-kunneman-vandenbosch:2013:WASSA}.

Early work on detecting sentiment focused on newswire text~\cite{Wiebe05,swn,Pang:2002:TUS,Hu04}.
As later research turned towards social media,
people realized this presented a number of new challenges.

Misspellings, poor grammatical structure, emoticons, acronyms, and slang were common in these new media,
%These characteristics of social media represent challenges that encourage new sentiment detection explorations.
and were explored by a number of researchers~\cite{Barbosa10,Bifet11,Davidov10,Jansen09,Kouloumpis11,oconnor10,Pak10}. Later, specialized shared tasks emerged, e.g., at SemEval \cite{Semeval2013,rosenthal-EtAl:2014:SemEval}, which compared teams against each other in a controlled environment using the same training and testing datasets. These shared tasks had the side effect to foster the emergence of a number of new resources, which eventually spread well beyond SemEval, e.g., NRC's Hashtag Sentiment lexicon and the Sentiment140 lexicon \cite{MohammadKZ2013}.\footnote{http://www.purl.com/net/lexicons}
%which were developed by the NRC team for their system for SemEval-2013 task 2, and were key elements in the success of their system, which was ranked first in the SemEval-2013 task.\footnote{http://www.purl.com/net/lexicons}

Below, we discuss the public evaluation done as part of SemEval-2015 Task 10. In its third year, the SemEval task on Sentiment Analysis in Twitter has once again attracted a large number of participants: 41 teams across five subtasks, with most teams participating in more than one subtask.

This year the task included reruns of two legacy subtasks,
which asked to detect the sentiment expressed in a tweet or by a particular phrase in a tweet.
The task further added three new subtasks. The first two focused on the sentiment towards a given topic in a single tweet or in a set of tweets, respectively.
%; detecting sentiment towards a topic is particularly useful for reviews where the target of the sentiment is important.
The third new subtask focused on determining the strength of prior association of Twitter terms with positive sentiment; this acts as an intrinsic evaluation of automatic methods that build Twitter-specific sentiment lexicons with real-valued sentiment association scores. 

%The Sentiment Analysis in Twitter Task consists of detecting polarity (as positive, negative, neutral/objective) in five subtasks: Subtask A evaluates the sentiment expressed by a phrase in the context of a tweet. Subtask B evaluates the overall sentiment of the tweet. Subtask C evaluates the sentiment towards the provided topic of the tweet. Subtask D evaluates the overall sentiment towards a topic across several tweets. Finally, in subtask E the objective is to determine the prior strength of association of Twitter terms (even hashtags and creatively spelled words) with positive sentiment. Even though named \emph{Sentiment Analysis in Twitter}, the tasks also includes evaluation on SMS and LiveJournal messages, as well as a special test set of sarcastic tweets.

In the remainder of this paper, we first introduce the problem of sentiment polarity classification and our subtasks. We then describe the process of creating the training, development, and testing datasets. We list and briefly describe the participating systems, the results, and the lessons learned. Finally, we compare the task to other related efforts and we point to possible directions for future research.

\section{Task Description}

Below, we describe the five subtasks of
SemEval-2015 Task 10 on Sentiment Analysis in Twitter.

\begin{itemize}[leftmargin=*]
%\subsection{Subtask A}

\item \textbf{Subtask A. Contextual Polarity Disambiguation:} Given an instance of a word/phrase in the context of a message, determine whether it expresses a positive, a negative or a neutral sentiment in that context.

%\subsection{Subtask B}

\item \textbf{Subtask B. Message Polarity Classification:} Given a message, determine whether it expresses a positive, a negative, or a neutral/objective sentiment. If both positive and negative sentiment are expressed, the stronger one should be chosen.

%\subsection{Subtask C}

\item \textbf{Subtask C. Topic-Based Message Polarity Classification:} Given a message and a topic, decide whether the message expresses a positive, a negative, or a neutral sentiment towards the topic. If both positive and negative sentiment are expressed, the stronger one should be chosen.

%\subsection{Subtask D}

\item \textbf{Subtask D. Detecting Trend Towards a Topic:} Given a set of messages on a given topic from the same period of time, classify the overall sentiment towards the topic in these messages as (a)~strongly positive, (b) weakly positive, (c) neutral, (d) weakly negative, or (e) strongly negative.

%\subsection{Subtask E}

%Many of the top performing sentiment analysis systems in recent SemEval competitions (2013 Task 2, 2014 Task 4, 2014 Task 9, and 2015 Task 10) rely on automatically generated sentiment lexicons. Sentiment lexicons are lists of words (and phrases) with prior associations to positive and negative sentiments. Some lexicons can additionally provide a sentiment score for a term to indicate its strength of evaluative intensity. Higher scores indicate greater intensity. Existing manually created sentiment lexicons tend to only have discrete labels for terms (positive, negative, neutral) but no real-valued scores indicating the intensity of sentiment. The goal of this task is to evaluate automatic methods of generating sentiment lexicons, especially those that also produce real-valued scores of sentiment intensity or association.

%\item \textbf{Subtask E. Determining Strength of Association of Twitter Terms with Positive Sentiment (Degree of Prior Polarity):} Given a word or a phrase, systems must provide a score between 0 and 1 that is indicative of its strength of association with positive sentiment. A score of 1 indicates maximum association with positive sentiment (or least association with negative sentiment) and a score of 0 indicates least association with positive sentiment (or maximum association with negative sentiment). If a word is more positive than another, then it should have a higher score than the other.

\item \textbf{Subtask E. Determining Strength of Association of Twitter Terms with Positive Sentiment (Degree of Prior Polarity):} Given a word/phrase, propose a score between 0 (lowest) and 1 (highest) that is indicative of the strength of association of that word/phrase with positive sentiment. If a word/phrase is more positive than another one, it should be assigned a relatively higher score.

\end{itemize}

\section{Datasets}

In this section, we describe the process of collecting and annotating our datasets
of short social media text messages.
We focus our discussion on the 2015 datasets;
more detail about the 2013 and the 2014 datasets can be found in \cite{Semeval2013} and \cite{rosenthal-EtAl:2014:SemEval}.

%We will focus our discussion on general tweets, but our testing data sets also include sarcastic tweets, SMS messages and sentences from LiveJournal, which we will also describe.

\subsection{Data Collection}

\begin{figure*}
\framebox{\parbox[t]{16.5cm}{\small\textbf{Instructions:} Subjective words are ones which convey an opinion or sentiment. Given a Twitter message, identify whether it is objective, positive, negative, or neutral. Then, identify each subjective word or phrase in the context of the sentence and mark the position of its start and end in the text boxes below. The number above each word indicates its position. The word/phrase will be generated in the adjacent textbox so that you can confirm that you chose the correct range. Choose the polarity of the word or phrase by selecting one of the radio buttons: positive, negative, or neutral. If a sentence is not subjective please select the checkbox indicating that ``There are no subjective words/phrases''. If a tweet is sarcastic, please select the checkbox indicating that ``The tweet is sarcastic''. Please read the examples and invalid responses before beginning if this is your first time answering this hit.}}
\framebox{\centering\parbox[t]{16.5cm}{\includegraphics[scale=.5]{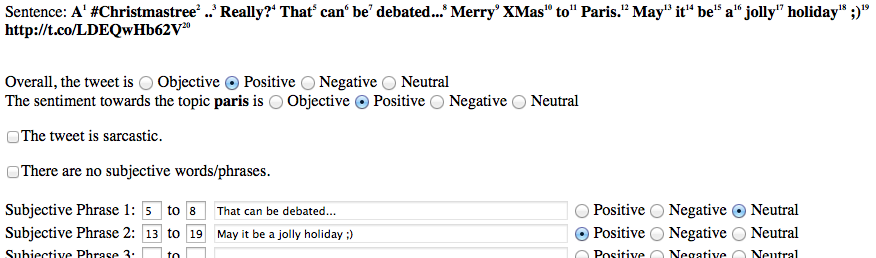}}}
\caption{The instructions we gave to the workers on Mechanical Turk, followed by a screenshot.}
\label{F:instructions}
\end{figure*}

\subsubsection{Subtasks A--D}

First, we gathered tweets that express sentiment about popular topics.
For this purpose, we extracted named entities from millions of tweets, using a Twitter-tuned NER system \cite{ner}. Our initial training set was collected over a one-year period spanning from January 2012 to January 2013. Each subsequent Twitter test set was collected a few months prior to the corresponding evaluation.
We used the public streaming Twitter API to download the tweets.
%\footnote{\url{https://dev.twitter.com/docs/streaming-apis}}

We then identified popular topics as those named entities that are frequently mentioned in association with a specific date \cite{ritter12}.
Given this set of automatically identified topics, we gathered tweets from the same time period which mentioned the named entities.
The testing messages had different topics from training and spanned later periods.

The collected tweets were greatly skewed towards the neutral class.
In order to reduce the class imbalance, we removed messages that contained no sentiment-bearing words using SentiWordNet as a repository of sentiment words.
Any word listed in SentiWordNet 3.0 with at least one sense having a positive or a negative sentiment score greater than 0.3 was considered a sentiment-bearing word.\footnote{Filtering based on an existing lexicon does bias the dataset to some degree; however, note that the text still contains sentiment expressions outside those in the lexicon.}

%We annotated the same Twitter messages for all subtasks. In contrast to prior years, 

For subtasks C and D, we did some manual pruning based on the topics. First, we excluded topics that were incomprehensible, ambiguous (e.g., \emph{Barcelona}, which is a name of a sports team and also of a place), or were too general (e.g., \emph{Paris}, which is a name of a big city). Second, we discarded tweets that were just mentioning the topic, but were not really about the topic. Finally, we discarded topics with too few tweets, namely less than 10.

\subsubsection{Subtask E}

\begin{table*}[t]
 \centering
\small
\begin{tabular}{|p{16cm}|}
\hline
 Authorities are \textit{\color{red}only too aware} that Kashgar is 4,000 kilometres (2,500 miles) from Beijing but \textit{\color{green}only} a tenth of the distance from the  Pakistani border, and are \textit{\color{red}desperate} to \textit{\color{red} ensure instability or militancy} does not leak over the frontiers. \\[3pt]
% \hline
Taiwan-made products \textit{\color{green} stood a good chance} of becoming \textit{\color{green} even more competitive thanks to} wider access to overseas markets and lower costs for material imports, he said. \\[3pt]
%\hline
``March \textit{\color{blue}appears} to be a \textit{\color{blue}more reasonable} estimate while earlier admission \textit{\color{blue}cannot be entirely ruled out},'' according to Chen, also Taiwan's chief WTO negotiator. \\[3pt]
%\hline
friday evening plans were great, but saturday's plans \textit{\color{red}didnt go as expected} -- i went dancing \& it was an \textit{\color{blue}ok} club, but \textit{\color{red}terribly crowded :-(} \\[3pt]
%\hline
WHY THE \textit{\color{red}HELL} DO YOU GUYS ALL HAVE MRS. KENNEDY! SHES A FUCKING DOUCHE \\[3pt]
%\hline
AT\&T was \textit{\color{blue}okay} but whenever they do something \textit{\color{green}nice} in the name of customer service it seems like a favor, while T-Mobile makes that a \textit{\color{green}normal everyday thin} \\[3pt]
%\hline
obama should be \textit{\color{red}impeached} on \textit{\color{red}TREASON} charges. Our Nuclear arsenal was TOP Secret. Till HE told our enemies what we had. \textit{\color{red}\#Coward \#Traitor} \\[3pt]
%\hline
My graduation speech: ``I'd like to \textit{\color{green}thanks} Google, Wikipedia and my computer!'' \textit{\color{green}:D} \#iThingteens \\
\hline
\end{tabular}
\caption{List of example sentences and annotations we provided to the Turkers. All subjective phrases are italicized and color-coded: positive phrases are in green, negative ones are in red, and neutral ones are in blue.}
\label{T:annotation-examples}
\end{table*}

\begin{table*}[t]
\centering
\small
%\begin{tabular}{@{}|@{ }@{ }l@{ }@{ }|@{ }@{ }c@{ }@{ }|@{}}
\begin{tabular}{@{}l@{ }@{ }@{ }@{ }|@{ }@{ }c@{}}
\hline
\textbf{\textit{I would love}} to watch Vampire Diaries \textbf{\textit{:)}} and some Heroes! \textbf{\textit{Great combination}} & 9/13 \\
I would love to watch Vampire Diaries :) and some \textbf{\textit{Heroes!}} \textbf{\textit{Great}} combination & 11/13 \\
I \textbf{\textit{would love}} to watch Vampire Diaries \textbf{\textit{:)}} and some Heroes! \textbf{\textit{Great}} combination & 10/13 \\
I would \textbf{\textit{love}} to watch Vampire Diaries :) and some Heroes! \textbf{\textit{Great}} combination & 13/13 \\
I would love to watch Vampire Diaries :) and some Heroes! \textbf{\textit{Great}} combination & 12/13\\
\hline
I would \textbf{\textit{love}} to watch Vampire Diaries :) and some Heroes! \textbf{\textit{Great}} combination & \\
\hline
\end{tabular}
\caption{Example of a sentence annotated for subjectivity on Mechanical Turk. Words and phrases that were marked as subjective are in bold italic. The first five rows are annotations provided by Turkers, and the final row shows their intersection. The last column shows the token-level accuracy for each annotation compared to the intersection.}
\label{T:Annotation}
\end{table*}

We selected high-frequency target terms from the Sentiment140 and the Hashtag Sentiment tweet corpora \cite{NRCJAIR14}.
In order to reduce the skewness towards the neutral class, we selected terms from different ranges of automatically determined sentiment values as provided by the corresponding Sentiment140 and Hashtag Sentiment lexicons.
The term set comprised regular English words, hashtagged words (e.g., \emph{\#loveumom}), misspelled or creatively spelled words (e.g., \emph{parlament} or \emph{happeeee}), abbreviations, shortenings, and slang. 
Some terms were negated expressions such as \emph{no fun}. (It is known that negation impacts the sentiment of its scope in complex ways \cite{zhu-EtAl:2014:P14-1}.) 
We annotated these terms for degree of sentiment manually. 
Further details about the data collection and the annotation process can be found in Section~\ref{sec:E:annotation} as well as in \cite{NRCJAIR14}.

The trial dataset consisted of 200 instances, and no training dataset was provided.
Note, however, that the trial data was large enough to be used as a development set,
or even as a training set.
Moreover, the participants were free to use any additional manually or automatically generated resources when building their systems for subtask E.
The testset included 1,315 instances.
%It could even be used for training. 
%(Note: the test data and the trial data had no terms in common.) 

\subsection{Annotation}

Below we describe the data annotation process.

\subsubsection{Subtasks A--D}

We used Amazon's Mechanical Turk for the annotations of subtasks A--D.
Each tweet message was annotated by five Mechanical Turk workers, also known as Turkers. The annotations for subtasks A--D were done concurrently, in a single task. A Turker had to mark all the subjective words/phrases in the tweet message by indicating their start and end positions and to say whether each subjective word/phrase was positive, negative, or neutral (subtask A). He/she also had to indicate the overall polarity of the tweet message in general (subtask B) as well as the overall polarity of the message towards the given target topic (subtasks C and D). The instructions we gave to the Turkers, along with an example,
are shown in Figure~\ref{F:instructions}.
We further made available to the Turkers several additional examples,
which we show in Table~\ref{T:annotation-examples}.

Providing all the required annotations for a given tweet message
%(a tweet, an SMS, or a sentence from LiveJournal)
constituted a Human Intelligence Task, or a HIT.
%Each Turker was paid 3-5 cents per hit.
In order to qualify to work on our HITs, a Turker had to have an approval rate greater than 95\% and should have completed at least 50 approved HITs.

We further discarded the following types of message annotations:

\begin{itemize}[noitemsep]
	\item containing overlapping subjective phrases;
	\item marked as subjective but having no annotated subjective phrases;
	\item with every single word marked as subjective;
	\item with no overall sentiment marked;
	\item with no topic sentiment marked.
\end{itemize}

\begin{table}[t]
\small
\begin{center}
\begin{tabular}{p{2.7cm} r r  r r}
\hline
\multicolumn{1}{c}{\bf Corpus} & \bf Pos. & \bf Neg. & \bf Obj. & \bf Total \\
&  &  & \bf / Neu. &\\
\hline
%Twitter Training  & 6,543 & 3,561 &  528  \\
Twitter2013-train  & 5,895 & 3,131 &  471 & 9,497\\
Twitter2013-dev  & 648 & 430 &  57 &  1,135 \\
Twitter2013-test  & 2,734 & 1,541 & 160 & 4,435\\
SMS2013-test  & 1,071 & 1,104 & 159 & 2,334\\
Twitter2014-test  & 1,807 & 578 & 88 & 2,473\\
\footnotesize{Twitter2014-sarcasm} & 82 & 37 & 5 & 124\\
\footnotesize{LiveJournal2014-test}  & 660 & 511 & 144 & 1,315\\
Twitter2015-test & 1899 & 1008 & 190 & 3097 \\
\hline
\end{tabular}
\caption{Dataset statistics for subtask A.}
\label{T:CorpusStatsA}
\end{center}
\end{table}

\begin{table}[t]
\small
\begin{center}
\begin{tabular}{p{2.7cm} r r  r r}
\hline
\multicolumn{1}{c}{\bf Corpus} & \bf Pos. & \bf Neg. & \bf Obj. & \bf Total\\
&  &  & \bf / Neu. &\\
\hline
%Twitter Training  & 4,115 & 1,720 &  5,047  \\
Twitter2013-train & 3,662 & 1,466 &  4,600 & 9,728\\
Twitter2013-dev & 575 & 340 &  739 & 1,654\\
Twitter2013-test  & 1,572 & 601 & 1,640 & 3,813\\
SMS2013-test  & 492 & 394 & 1,207 & 2,093\\
Twitter2014-test  & 982 & 202 & 669 & 1,853\\
\footnotesize{Twitter2014-sarcasm} & 33 & 40 & 13 & 86\\
\footnotesize{LiveJournal2014-test}  & 427 & 304 & 411 & 1,142\\
Twitter2015-test & 1040 & 365 & 987 & 2392 \\
\hline
\end{tabular}
\caption{Dataset statistics for subtask B.}
\label{T:CorpusStatsB}
\end{center}
\end{table}

\begin{table}[t]
\small
\begin{center}
\begin{tabular}{l r r r r r}
\hline
\multicolumn{1}{c}{\bf Corpus} & \bf Topics & \bf Pos. & \bf Neg. & \bf Obj. & \bf Total\\
 &  & &  & \bf / Neu. & \\
\hline
%Twitter Training  & 4,115 & 1,720 &  5,047  \\
Train & 44 & 142 & 56 & 288 & 530 \\
Test & 137 & 870 & 260 & 1256 & 2386 \\
\hline
\end{tabular}
\caption{Twitter-2015 statistics for subtasks C \& D.}
\label{T:CorpusStatsC}
\end{center}
\end{table}

Recall that each tweet message was annotated by five different Turkers.
We consolidated these annotations for subtask A 
using intersection as shown in the last row of Table~\ref{T:Annotation}.
A word had to appear in 3/5 of the annotations in order to be considered subjective.
It further had to be labeled with a particular polarity (positive, negative, or neutral) by three of the five Turkers in order to receive that polarity label.
As the example shows, this effectively shortens the spans of the annotated phrases,
often to single words, as it is hard to agree on long phrases.

We also experimented with two alternative methods for combining annotations:
({\it i})~by computing the union of the annotations for the sentence,
and ({\it ii})~by taking the annotations by the Turker who has annotated the highest number of HITs.
However, our manual analysis has shown that both alternatives performed worse than using the intersection.
%so eventually we only used intersection.

For subtasks B and C, we consolidated the tweet-level annotations using majority voting,
requiring that the winning label be proposed by at least three of the five Turkers;
we discarded all tweets for which 3/5 majority could not be achieved.
As in previous years,
%in order to reduce the number of rejected sentences,
we combined the objective and the neutral labels, which Turkers tended to mix up. 

%For the sarcastic tweets, we slightly altered the annotation task. The tweets were shown to the Turkers without the \#sarcasm hashtag,
%and the Turkers were asked to determine whether the tweet is sarcastic on its own.
%Furthermore, the Turkers had to indicate the degree of sarcasm as
%(a)~definitely sarcastic, (b)~probably sarcastic, and (c)~not sarcastic. Although we do not use the degree of sarcasm at this time, it could be useful for analysis as well as possibly excluding tweets that do not appear to be sarcastic.
%TODO: \textbf{Svetlana: What did you do with these annotations for sarcasm?}
%For the SMS and the LiveJournal messages, the annotation task was the same as for tweets.
%We further collected a new test dataset for SemEval-2014 Task 9.

\begin{table*}[t]
\small
\begin{center}
\begin{tabular}{l p{11cm} c}
\hline
\multicolumn{1}{l}{\bf Source} & \multicolumn{1}{c}{\bf Message} & \bf Message-Level \\
& & \bf Polarity \\
\hline
Twitter & Why would you [still]- wear shorts when it's this cold?!  I [love]+ how Britain see's a bit of sun and they're [like 'OOOH]+ LET'S STRIP!' & positive \\
%\hline
SMS & [Sorry]- I think tonight [cannot]- and I [not feeling well]- after my rest. & negative \\
%\hline
LiveJournal & [Cool]+ posts , dude ; very [colorful]+ , and [artsy]+ . & positive \\
%\hline
Twitter Sarcasm & [Thanks]+ manager for putting me on the schedule for Sunday & negative \\
\hline
\end{tabular}
\caption{Example annotations for each source of messages.
        The subjective phrases are marked in [$\ldots$], and are followed by their polarity (subtask A);
        the message-level polarity is shown in the last column (subtask B).}
\label{T:Examples}
\end{center}
\end{table*}

\begin{table*}[t]
\small
\begin{center}
\begin{tabular}{l p{9cm} c c}
\hline
\multicolumn{1}{l}{\bf Topic} & \multicolumn{1}{c}{\bf Message} & \bf Message-Level & \bf Topic-Level \\
& & \bf Polarity & \bf Polarity \\
\hline
leeds united & Saturday without Leeds United is like Sunday dinner it doesn't feel normal at all (Ryan) & negative & positive \\
%\hline
demi lovato & Who are you tomorrow? Will you make me smile or just bring me sorrow? \#HottieOfTheWeek Demi Lovato & neutral & positive \\
\hline
\end{tabular}
\caption{Example of annotations in Twitter showing differences between topic- and message-level polarity.}
\label{T:B-CExamples}
\end{center}
\end{table*}

We used these consolidated annotations as gold labels for subtasks A, B, C \& D.
%consecutive tokens marked as subjective served as target terms for subtask A.
The statistics for all datasets for these subtasks are shown in Tables~\ref{T:CorpusStatsA}, \ref{T:CorpusStatsB}, and \ref{T:CorpusStatsC}, respectively.
Each dataset is marked with the year of the SemEval edition it was produced for.
An annotated example from each source (Twitter, SMS, LiveJournal)
is shown in Table~\ref{T:Examples};
examples for sentiment towards a topic can be seen in Table~\ref{T:B-CExamples}.

%For subtask D, we did not use the Mechanical Turk, but we did the annotations ourselves:
%each set of tweets for a given topic was annotated by two different task organizers,
%and we resolved disagreements by discussion.

%The final testing datasets for 2015 overlap only partially between the subtasks since we had to discard messages with low inter-annotator agreement, and this differed between the subtasks.
%
%After the annotation process, we split the annotated tweets into training, development and testing;
%for testing, we further annotated three additional out-of-domain datasets:
%
%\begin{itemize}
%    \item \textbf{SMS messages:} from the NUS SMS corpus\footnote{\texttt{http://wing.comp.nus.edu.sg/SMSCorpus/}} \cite{SMScorpus};
%    \item \textbf{LiveJournal:} sentences from LiveJournal;
%    \item \textbf{Sarcastic tweets:} a small set of tweets containing the \#sarcasm hashtag.
%\end{itemize}

\subsubsection{Subtask E}
\label{sec:E:annotation}

Subtask E asks systems to propose a numerical score
for the positiveness of a given word or phrase.
Many studies have shown that people are actually quite bad at assigning such absolute scores:
inter-annotator agreement is low, and annotators struggle even to remain self-consistent.
%different people may assign very different scores for the same example,
%nd it is often hard for annotators to remain even self-consistent.
In contrast, it is much easier to make relative judgments,
e.g., to say whether one word is more positive than another.
Moreover, it is possible to derive an absolute score from pairwise judgments, but this requires a much larger number of annotations. Fortunately, there are schemes that allow to infer more pairwise annotations from less judgments.

One such annotation scheme is MaxDiff \cite{Louviere91},
%that retains the comparative aspect while still requiring only a small number of annotations \cite{Louviere91}.
which is widely used in market surveys \cite{Almquist_2009}; it was also used in a previous SemEval task
\cite{jurgens12}.
%for determining relation similarity of pairs of items by \newcite{jurgens12} in a SemEval-2012 task.

In MaxDiff, the annotator is presented with four terms and asked
which term is most positive and which is least positive.
By answering just these two questions,
five out of six pairwise rankings become known.
Consider a set in which a judge evaluates $A$, $B$, $C$, and $D$.
If she says that $A$ and $D$ are the most and the least positive,
%respectively,
%then
we can infer the following:
%pairwise rankings:
$A > B, A > C, A > D, B > D, C > D$.
The responses to the MaxDiff questions can then be easily translated into a ranking for all the terms and also into a real-valued score for each term.  
We crowdsourced the MaxDiff questions on CrowdFlower, recruiting ten annotators per MaxDiff example.
Further details can be found in Section  6.1.2. of \cite{NRCJAIR14}.

\subsection{Lower \& Upper Bounds}

When building a system to solve a task, it is good to know how well we should expect it to perform. One good reference point is agreement between annotators. Unfortunately, as we derive annotations by agreement, we cannot calculate standard statistics such as Kappa. Instead, we decided to measure the agreement between our gold standard annotations (derived by agreement) and the annotations proposed by the best Turker, the worst Turker, and the average Turker (with respect to the gold/consensus annotation for a particular message). Given a HIT, we just calculate the overlaps as shown in the last column in Table~\ref{T:Annotation}, and then we calculate the best, the worst, and the average, which are respectively 13/13, 9/13 and 11/13, in the example. Finally, we average these statistics over all HITs that contributed to a given dataset, to produce lower, average, and upper averages for that dataset.
The accuracy (with respect to the gold/consensus annotation) for different averages is shown in Table~\ref{T:bounds}. Since the overall polarity of a message is chosen based on majority, the upper bound for subtask B is 100\%. These averages give a good indication about how well we can expect the systems to perform. We can see that even if we used the best annotator for each HIT, it would still not be possible to get perfect accuracy, and thus we should also not expect it from a system.

%\textbf{**** SARA: Preslav, is there a way to easily compute the ratios for subtask D using the scorer? }

\begin{table}[h]
\small
\begin{center}
\begin{tabular}{ l  c  c  c  c }
\hline
\bf Corpus & \multicolumn{3}{c}{\bf Subtask A} & \multicolumn{1}{c}{\bf Subtask B} \\
%\cline{2-4}
& \bf Low & \bf Avg & \bf Up & \multicolumn{1}{c}{\bf Avg} \\
\hline
Twitter2013-train & 75.1 & 89.7 & 97.9 & 77.6 \\
Twitter2013-dev & 66.6 & 85.3 & 97.1 & 86.4 \\
Twitter2013-test & 76.8 & 90.3 & 98.0 & 75.9 \\
SMS2013-test & 75.9 & 97.5 & 89.6 & 77.5 \\
\footnotesize{Livejournal2014-test} & 61.7 & 82.3 & 94.5 & 76.2 \\
Twitter2014-test & 75.3 & 88.9 & 97.5 & 74.7 \\
Sarcasm2014-test & 62.6 & 83.1 & 95.6 & 71.2 \\
Twitter2015-test & 73.2 & 87.6 & 96.8 & 75.7 \\
\hline
\end{tabular}
\caption{Average (over all HITs) overlap of the gold annotations with the worst, average, and the worst Turker for each HIT, for subtasks A and B.}
%\caption{The lower (low), average (avg), and upper (up) bounds for the datasets in subtasks A and B computed using the answers of individual Mechanical Turk workers.}
\label{T:bounds}
\end{center}
\end{table}

\subsection{Tweets Delivery}

Due to restrictions in the Twitter's terms of service, we could not deliver the annotated tweets to the participants directly.
Instead, we released annotation indexes and labels, a list of corresponding Twitter IDs, and a download script that extracts the corresponding tweets via the Twitter API.\footnote{https://dev.twitter.com}

As a result, different teams had access to different number of training tweets depending on when they did the downloading.
However, our analysis has shown that this did not have a major impact and many high-scoring teams had less training data compared to some lower-scoring ones.

\section{Scoring}
\label{Section:scoring}

\subsection{Subtasks A-C: Phrase-Level, Message-Level, and Topic-Level Polarity}

The participating systems were required to perform a three-way classification,
i.e., to assign one of the folowing three labels:
{\em positive}, {\em negative} or {\em objective/neutral}.
We evaluated the systems in terms of a macro-averaged $F_1$ score for predicting positive and negative phrases/messages.

We first computed positive precision, $P_{pos}$ as follows:
we found the number of phrases/messages that a system correctly predicted to be positive,
and we divided that number by the total number of examples it predicted to be positive.
To compute positive recall, $R_{pos}$, we found the number of phrases/messages correctly predicted to be positive and we divided that number by the total number of positives in the gold standard.
We then calculated an $F_1$ score for the positive class
as follows $F_{pos}=\frac{2 P_{pos} R_{pos}}{P_{pos} + R_{pos}}$. 
We carried out similar computations for the negative phrases/messages, $F_{neg}$.
The overall score
%for each system run
was then computed as the average of the $F_1$scores for the positive and for the negative classes:
$F=(F_{pos}+F_{neg})/2$.

We provided the participants with a scorer that outputs the overall score $F$,
as well as $P$, $R$, and $F_1$ scores for each class (positive, negative, neutral) and for each test set.

\subsection{Subtask D: Overall Polarity Towards a Topic}

This subtask asks to predict the overall sentiment of a set of tweets towards a given topic.
In other words, to predict the ratio $r_i$ of positive ($pos_i$) tweets
to the number of positive and negative sentiment tweets in the set of tweets
about the $i$-th topic:
$$r_i = Pos_i/(Pos_i+Neg_i)$$

Note, that neutral tweets do not participate in the above formula;
they have only an indirect impact on the calculation,
similarly to subtasks A--C.

We use the following two evaluation measures for subtask D:

\begin{itemize}
\item \textbf{AvgDiff} (official score): Calculates the absolute difference betweeen the predicted $r^\prime_i$ and the gold $r_i$ for each $i$, and then averages this difference over all topics.
\item \textbf{AvgLevelDiff} (unofficial score): This calculation is the same as AvgDiff, but with $r^\prime_i$ and $r_i$ first remapped to five coarse numerical categories: 5 (strongly positive), 4 (weakly positive), 3 (mixed), 2 (weakly negative), and 1 (strongly negative). We define this remapping based on intervals as follows:
\begin{itemize}
\item 5: $0.8 < x \le 1.0$
\item 4: $0.6 < x \le 0.8$
\item 3: $0.4 < x \le 0.6$
\item 2: $0.2 < x \le 0.4$
\item 1: $0.0 \le x \le 0.2$
\end{itemize}
\end{itemize}

\subsection{Subtask E: Degree of Prior Polarity}

The scores proposed by the participating systems were evaluated by first ranking the terms according to the proposed sentiment score and then comparing this ranked list to a ranked list obtained from aggregating the human ranking annotations. We used Kendall's rank correlation  %coefficient
(Kendall's $\tau$) as the official evaluation metric to compare the ranked lists \cite{kendall1938new}. We also calculated scores for Spearman's rank correlation \cite{Spearman},
as an unofficial score.

\section{Participants and Results}

\begin{table}[t]
\begin{center}
\tiny{
\begin{tabular}{ll}
\bf Team ID & \bf Affiliation\\
\hline
CIS-positiv & University of Munich\\
CLaC-SentiPipe & CLaC Labs, Concordia University\\
DIEGOLab & Arizona State University\\
ECNU & East China Normal University\\
elirf & Universitat Polit\`{e}cnica de Val\`{e}ncia\\
Frisbee & Frisbee\\
Gradiant-Analytics & Gradiant\\
GTI & AtlantTIC Center, University of Vigo\\
IHS-RD & IHS inc\\
iitpsemeval & Indian Institute of Technology, Patna\\
IIIT-H & IIIT, Hyderabad\\
INESC-ID & IST, INESC-ID\\
IOA & Institute of Acoustics, Chinese Academy of Sciences\\
KLUEless & FAU Erlangen-N\"{u}rnberg\\
lsislif & Aix-Marseille University\\
NLP & NLP\\
RGUSentimentMiners123 & Robert Gordon University\\
RoseMerry & The University of Melbourne\\
Sentibase & IIIT, Hyderabad\\
SeNTU & Nanyang Technological University, Singapore\\
SHELLFBK & Fondazione Bruno Kessler\\
sigma2320 & Peking University\\
Splusplus & Beihang University\\
SWASH & Swarthmore College\\
SWATAC & Swarthmore College\\
SWATCMW & Swarthmore College\\
SWATCS65 & Swarthmore College\\
Swiss-Chocolate & Zurich University of Applied Sciences\\
TwitterHawk & University of Massachusetts, Lowell\\
UDLAP2014 & Universidad de las Am\`{e}ricas Puebla, Mexico\\
UIR-PKU & University of International Relations\\
UMDuluth-CS8761 & University of Minnesota, Duluth\\
UNIBA & University of Bari Aldo Moro\\
unitn & University of Trento\\
UPF-taln & Universitat Pompeu Fabra\\
WarwickDCS & University of Warwick\\
Webis & Bauhaus-Universit\"{a}t Weimar\\
whu-iss & International Software School, Wuhan University\\
Whu-Nlp & Computer School, Wuhan University\\
wxiaoac & Hong Kong University of Science and Technology\\
ZWJYYC & Peking University\\
\hline
\end{tabular}
}
\caption{The participating teams and their affiliations.}
\label{tab:participants}
\end{center}
\vspace*{-5mm}
\end{table}

The task attracted 41 teams:
11 teams participated in subtask A,
40 in subtask B,
 7 in subtask C,
 6 in subtask D, and
10 in subtask E.
The IDs and affiliations of the participating teams are shown in Table~\ref{tab:participants}.

\subsection{Subtask A: Phrase-Level Polarity}

\begin{table*}[t!]
    \centering
    \begin{small}
    \begin{tabular}{c|l|cc|ccc|c}
    \hline
   & & \multicolumn{2}{c|}{\bf 2013: Progress} & \multicolumn{3}{c|}{\bf 2014: Progress} & \bf 2015: Official \\
   \bf \# & \multicolumn{1}{|c|}{\bf System} & \bf Tweet & \bf SMS & \bf Tweet & \bf Tweet & \bf Live- & \bf Tweet \\
   & &  &  & \bf & \bf sarcasm & \bf Journal & \\
    \hline
	1 & unitn & 90.10$_{1}$ & 88.60$_{2}$ & 87.12$_{1}$ & 73.65$_{5}$ & 84.46$_{2}$ & \bf 84.79$_{1}$\\
	2 & KLUEless & 88.56$_{2}$ & 88.62$_{1}$ & 84.99$_{3}$ & 75.59$_{4}$ & 83.94$_{4}$ & \bf 84.51$_{2}$\\
	3 & IOA & 83.90$_{7}$ & 84.18$_{7}$ & 85.37$_{2}$ & 71.58$_{6}$ & 85.61$_{1}$ & \bf 82.76$_{3}$\\
	4 & WarwickDCS & 84.08$_{6}$ & 84.40$_{5}$ & 83.89$_{5}$ & 78.03$_{2}$ & 83.18$_{5}$ & \bf 82.46$_{4}$\\
	5 & TwitterHawk & 82.87$_{8}$ & 83.64$_{8}$ & 84.05$_{4}$ & 75.62$_{3}$ & 83.97$_{3}$ & \bf 82.32$_{5}$\\
	6 & iitpsemeval & 85.81$_{3}$ & 85.86$_{3}$ & 82.73$_{6}$ & 65.71$_{9}$ & 81.76$_{7}$ & \bf 81.31$_{6}$\\
	7 & ECNU & 85.28$_{4}$ & 84.70$_{4}$ & 82.09$_{7}$ & 70.96$_{7}$ & 82.49$_{6}$ & \bf 81.08$_{7}$\\
	8 & Whu-Nlp & 79.76$_{9}$ & 81.78$_{9}$ & 81.69$_{8}$ & 63.14$_{11}$ & 80.87$_{9}$ & \bf 78.84$_{8}$\\
	9 & GTI & 84.64$_{5}$ & 84.37$_{6}$ & 79.48$_{9}$ & 81.53$_{1}$ & 81.61$_{8}$ & \bf 77.27$_{9}$\\
	10 & whu-iss & 74.02$_{10}$ & 70.26$_{11}$ & 72.20$_{10}$ & 69.33$_{8}$ & 73.57$_{10}$ & \bf 71.35$_{10}$\\
	11 & UMDuluth-CS8761 & 72.71$_{11}$ & 71.80$_{10}$ & 69.84$_{11}$ & 64.53$_{10}$ & 71.53$_{11}$ & \bf 66.21$_{11}$\\
	\hline
	& baseline & 38.1 & 31.5 & 42.2 & 39.8 & 33.4 & \bf 38.0\\
	\hline
    \end{tabular}
    \end{small}
    \caption{\label{tab:res-subtaskA} \textbf{Results for subtask A: Phrase-Level Polarity.}
             The systems are ordered by their score on the Twitter2015 test dataset;
             the rankings on the individual datasets are indicated with a subscript.}
  \end{table*}

The results (macro-averaged $F_1$ score) for subtask A are shown in Table~\ref{tab:res-subtaskA}.
The official results on the new Twitter2015-test dataset are shown in the last column,
while the first five columns show $F_1$ on the 2013 and on the 2014 progress test
datasets:\footnote{Note that the 2013 and the 2014 test datasets were made available for development, but it was explicitly forbidden to use them for training.}
Twitter2013-test, SMS2013-test, Twitter2014-test, Twitter2014-sarcasm, and LiveJournal2014-test.
There is an index for each result showing the relative rank of that result within the respective column. The participating systems are ranked by their score on the Twitter2015-test dataset, which is the official ranking for subtask A; all remaining rankings are secondary.

There were less participants this year,
% compared to previous years,
probably due to having a new similar subtask: C.
% and thus compete against A.
Notably, many of the participating teams were newcomers.

We can see that all systems beat the majority class baseline by 25-40
$F_1$ points absolute on all datasets.
The winning team unitn (using deep convolutional neural networks)
achieved an $F_1$ of 84.79 on Twitter2015-test,
followed closely by KLUEless (using logistic regression) with $F_1$=84.51.

Looking at the progress datasets, we can see that
unitn was also first on both progress Tweet datasets,
and second on SMS and on LiveJournal.
KLUEless won SMS and was second on Twitter2013-test.
The best result on LiveJournal was achieved by IOA, 
who were also second on Twitter2014-test and third on the official Twitter2015-test.
None of these teams was ranked in top-3 on Twitter2014-sarcasm,
where the best team was GTI, followed by WarwickDCS.

Compared to 2014, there is an improvement on Twitter2014-test
from 86.63 in 2014 (NRC-Canada) to 87.12 in 2015 (unitn).
The best result on Twitter2013-test of 90.10 (unitn) this year
is very close to the best in 2014 (90.14 by NRC-Canada).
Similarly, the best result on LiveJournal stays exactly the same,
i.e., $F_1$=85.61 (SentiKLUE in 2014 and IOA in 2015).
However, there is slight degradation for SMS2013-test
from 89.31 (ECNU) in 2014 to 88.62 (KLUEless) in 2015.
The results also degraded for Twitter2014-sarcasm
from 82.75 (senti.ue) to 81.53 (GTI).

\subsection{Subtask B: Message-Level Polarity}

\begin{table*}[t!]
    \centering
    \begin{small}
    \begin{tabular}{c|l|cc|ccc|cc}
    \hline
   & & \multicolumn{2}{c|}{\bf 2013: Progress} & \multicolumn{3}{c|}{\bf 2014: Progress} & \multicolumn{2}{c}{\bf 2015: Official} \\
   \bf \# & \multicolumn{1}{|c|}{\bf System} & \bf Tweet & \bf SMS & \bf Tweet & \bf Tweet & \bf Live- & \bf Tweet & \bf Tweet \\
   & &  &  & \bf & \bf sarcasm & \bf Journal & & \bf sarcasm\\
    \hline
	1 & Webis & 68.49$_{10}$ & 63.92$_{14}$ & 70.86$_{7}$ & 49.33$_{12}$ & 71.64$_{14}$ & \bf 64.84$_{1}$ & 53.59$_{22}$\\
	2 & unitn & 72.79$_{2}$ & 68.37$_{2}$ & 73.60$_{2}$ & 55.44$_{5}$ & 72.48$_{12}$ & \bf 64.59$_{2}$ & 55.01$_{19}$\\
	3 & lsislif & 71.34$_{4}$ & 63.42$_{17}$ & 71.54$_{5}$ & 46.57$_{22}$ & 73.01$_{10}$ & \bf 64.27$_{3}$ & 46.00$_{33}$\\
	4 & INESC-ID$^\star$ & 71.97$_{3}$ & 63.78$_{15}$ & 72.52$_{3}$ & 56.23$_{3}$ & 69.78$_{22}$ & \bf 64.17$_{4}$ & 64.91$_{2}$\\
	5 & Splusplus & 72.80$_{1}$ & 67.16$_{5}$ & 74.42$_{1}$ & 42.86$_{31}$ & 75.34$_{1}$ & \bf 63.73$_{5}$ & 60.99$_{7}$\\
	6 & wxiaoac & 66.43$_{16}$ & 64.04$_{13}$ & 68.96$_{11}$ & 54.38$_{7}$ & 73.36$_{9}$ & \bf 63.00$_{6}$ & 52.22$_{26}$\\
	7 & IOA & 71.32$_{5}$ & 68.14$_{3}$ & 71.86$_{4}$ & 51.48$_{9}$ & 74.52$_{2}$ & \bf 62.62$_{7}$ & 65.77$_{1}$\\
	8 & Swiss-Chocolate & 68.80$_{9}$ & 65.56$_{6}$ & 68.74$_{12}$ & 48.22$_{16}$ & 73.95$_{4}$ & \bf 62.61$_{8}$ & 54.66$_{20}$\\
	9 & CLaC-SentiPipe & 70.42$_{7}$ & 63.05$_{18}$ & 70.16$_{10}$ & 51.43$_{10}$ & 73.59$_{6}$ & \bf 62.00$_{9}$ & 58.55$_{9}$\\
	10 & TwitterHawk & 68.44$_{11}$ & 62.12$_{20}$ & 70.64$_{9}$ & 56.02$_{4}$ & 70.17$_{19}$ & \bf 61.99$_{10}$ & 61.24$_{6}$\\
	11 & SWATCS65 & 68.21$_{12}$ & 65.49$_{8}$ & 67.23$_{14}$ & 37.23$_{39}$ & 73.37$_{8}$ & \bf 61.89$_{11}$ & 52.64$_{24}$\\
	12 & UNIBA & 61.66$_{29}$ & 65.50$_{7}$ & 65.11$_{25}$ & 37.30$_{38}$ & 70.05$_{20}$ & \bf 61.55$_{12}$ & 48.16$_{32}$\\
	13 & KLUEless & 70.64$_{6}$ & 67.66$_{4}$ & 70.89$_{6}$ & 45.36$_{26}$ & 73.50$_{7}$ & \bf 61.20$_{13}$ & 56.19$_{17}$\\
	14 & NLP & 66.96$_{14}$ & 61.05$_{25}$ & 67.45$_{13}$ & 39.87$_{34}$ & 66.12$_{31}$ & \bf 60.93$_{14}$ & 63.62$_{3}$\\
	15 & ZWJYYC & 69.56$_{8}$ & 64.72$_{11}$ & 70.77$_{8}$ & 46.34$_{23}$ & 71.60$_{15}$ & \bf 60.77$_{15}$ & 52.40$_{25}$\\
	16 & Gradiant-Analytics & 65.29$_{22}$ & 61.97$_{21}$ & 66.87$_{17}$ & 59.11$_{1}$ & 72.63$_{11}$ & \bf 60.62$_{16}$ & 56.45$_{16}$\\
	17 & IIIT-H & 65.68$_{20}$ & 62.25$_{19}$ & 67.04$_{16}$ & 57.50$_{2}$ & 69.91$_{21}$ & \bf 59.83$_{17}$ & 62.75$_{5}$\\
	18 & ECNU & 65.25$_{23}$ & 68.49$_{1}$ & 66.37$_{20}$ & 45.87$_{25}$ & 74.40$_{3}$ & \bf 59.72$_{18}$ & 52.67$_{23}$\\
	19 & CIS-positiv & 64.82$_{24}$ & 65.14$_{10}$ & 66.05$_{21}$ & 49.23$_{14}$ & 71.47$_{16}$ & \bf 59.57$_{19}$ & 57.74$_{11}$\\
	20 & SWASH & 63.07$_{27}$ & 56.49$_{34}$ & 62.93$_{31}$ & 48.42$_{15}$ & 69.43$_{24}$ & \bf 59.26$_{20}$ & 54.30$_{21}$\\
	21 & GTI & 64.03$_{25}$ & 63.50$_{16}$ & 65.65$_{22}$ & 55.38$_{6}$ & 70.50$_{17}$ & \bf 58.95$_{21}$ & 57.02$_{13}$\\
	22 & iitpsemeval & 60.78$_{31}$ & 60.56$_{26}$ & 65.09$_{26}$ & 47.32$_{19}$ & 73.70$_{5}$ & \bf 58.80$_{22}$ & 58.18$_{10}$\\
	23 & elirf & 57.05$_{32}$ & 60.20$_{28}$ & 61.17$_{35}$ & 45.98$_{24}$ & 68.33$_{28}$ & \bf 58.58$_{23}$ & 43.91$_{34}$\\
	24 & SWATAC & 65.86$_{19}$ & 61.30$_{24}$ & 66.64$_{19}$ & 39.45$_{35}$ & 68.67$_{27}$ & \bf 58.43$_{24}$ & 50.66$_{27}$\\
	25 & UIR-PKU$^\star$ & 67.41$_{13}$ & 64.67$_{12}$ & 67.18$_{15}$ & 52.58$_{8}$ & 70.44$_{18}$ & \bf 57.65$_{25}$ & 59.43$_{8}$\\
	26 & SWATCMW & 65.67$_{21}$ & 65.43$_{9}$ & 65.62$_{23}$ & 37.48$_{36}$ & 69.52$_{23}$ & \bf 57.60$_{26}$ & 56.69$_{14}$\\
	27 & WarwickDCS & 66.57$_{15}$ & 61.92$_{22}$ & 65.47$_{24}$ & 45.03$_{28}$ & 68.98$_{25}$ & \bf 57.32$_{27}$ & 56.58$_{15}$\\
	28 & SeNTU & 63.50$_{26}$ & 60.53$_{27}$ & 66.85$_{18}$ & 45.18$_{27}$ & 68.70$_{26}$ & \bf 57.06$_{28}$ & 49.53$_{29}$\\
	29 & DIEGOLab & 62.49$_{28}$ & 58.60$_{30}$ & 63.99$_{28}$ & 47.62$_{18}$ & 63.74$_{34}$ & \bf 56.72$_{29}$ & 55.56$_{18}$\\
	30 & Sentibase & 61.56$_{30}$ & 59.26$_{29}$ & 63.29$_{30}$ & 47.07$_{20}$ & 67.55$_{29}$ & \bf 56.67$_{30}$ & 62.96$_{4}$\\
	31 & Whu-Nlp & 65.97$_{18}$ & 61.31$_{23}$ & 63.93$_{29}$ & 46.93$_{21}$ & 71.83$_{13}$ & \bf 56.39$_{31}$ & 22.25$_{40}$\\
	32 & UPF-taln & 66.15$_{17}$ & 57.84$_{31}$ & 65.05$_{27}$ & 50.93$_{11}$ & 64.50$_{32}$ & \bf 55.59$_{32}$ & 41.63$_{35}$\\
	33 & RGUSentimentMiners123 & 56.41$_{34}$ & 57.14$_{32}$ & 59.44$_{36}$ & 44.72$_{29}$ & 64.39$_{33}$ & \bf 53.73$_{33}$ & 48.21$_{31}$\\
	34 & IHS-RD$^\star$ & 55.06$_{35}$ & 57.08$_{33}$ & 61.39$_{32}$ & 37.32$_{37}$ & 66.99$_{30}$ & \bf 52.65$_{34}$ & 36.02$_{37}$\\
	35 & RoseMerry & 52.33$_{37}$ & 53.00$_{36}$ & 61.27$_{34}$ & 49.25$_{13}$ & 62.54$_{35}$ & \bf 51.18$_{35}$ & 49.62$_{28}$\\
	36 & Frisbee & 49.37$_{38}$ & 46.59$_{38}$ & 53.92$_{38}$ & 42.07$_{32}$ & 57.94$_{38}$ & \bf 49.19$_{36}$ & 48.26$_{30}$\\
	37 & UMDuluth-CS8761 & 54.17$_{36}$ & 50.64$_{37}$ & 55.82$_{37}$ & 43.74$_{30}$ & 60.23$_{37}$ & \bf 47.77$_{37}$ & 34.40$_{38}$\\
	38 & UDLAP2014 & 41.93$_{39}$ & 39.35$_{39}$ & 45.93$_{39}$ & 41.04$_{33}$ & 50.11$_{39}$ & \bf 42.10$_{38}$ & 40.59$_{36}$\\
	39 & SHELLFBK & 32.14$_{40}$ & 26.14$_{40}$ & 32.20$_{40}$ & 35.58$_{40}$ & 34.06$_{40}$ & \bf 32.45$_{39}$ & 25.73$_{39}$\\
	40 & whu-iss & 56.51$_{33}$ & 54.28$_{35}$ & 61.31$_{33}$ & 47.78$_{17}$ & 61.98$_{36}$ & \bf 24.80$_{40}$ & 57.73$_{12}$\\
	\hline
	   & baseline & 29.2 & 19.0 & 34.6 & 27.7 & 27.2 & \bf 30.3 & 30.2 \\
	\hline
    \end{tabular}
    \end{small}
    \caption{\label{tab:res-subtaskB} \textbf{Results for subtask B: Message-Level Polarity.}
             The systems are ordered by their score on the Twitter2015 test dataset;
             the rankings on the individual datasets are indicated with a subscript.
             Systems with late submissions for the \emph{progress} test datasets
             (but with timely submissions for the official 2015 test dataset) are marked with a $^\star$.}
  \end{table*}

The results for subtask B are shown in Table~\ref{tab:res-subtaskB}.
Again, we show results on the five progress test datasets from 2013 and 2014,
in addition to those for the official Twitter2015-test datasets.

Subtask B attracted 40 teams, 
both newcomers and returning,
similarly to 2013 and 2014.
All managed to beat the baseline
with the exception of one system for Twitter2015-test,
and one for Twitter2014-test.
There is a cluster of four teams at the top:
Webis (ensemble combining four Twitter sentiment classification approaches
that participated in previous editions) with an $F_1$ of 64.84,
unitn with 64.59,
lsislif (logistic regression with special weighting for positives and negatives) with 64.27,
and
INESC-ID (word embeddings) with 64.17.

The last column in the table shows the results for the 2015 sarcastic tweets. Note that, unlike in 2014, this time they were not collected separately and did not have a special \#sarcasm tag; instead, they are a subset of 75 tweets from Twitter2015-test that were flagged as sarcastic by the human annotators. The top system is
IOA with an $F_1$ of 65.77,
followed by INESC-ID with 64.91,
and NLP with 63.62.

Looking at the progress datasets,
we can see that the second ranked unitn is also second on SMS and on Twitter2014-test,
and third on Twitter2013-test.
INESC-ID in turn is third on Twitter2014-test and also third on Twitter2014-sarcasm.
Webis and lsislif were less strong on the progress datasets.

Compared to 2014,
there is improvement on Twitter2013-test from 72.12 (TeamX) to 72.80 (Splusplus),
on Twitter2014-test from 70.96 (TeamX) to 74.42 (Spluplus),
on Twitter2014-sarcasm from 58.16 (NRC-Canada) to 59.11 (Gradiant-Analytics),
and on LiveJournal from 74.84 (NRC-Canada) to 75.34 (Splusplus),
but not on SMS: 70.28 (NRC-Canada) vs. 68.49 (ECNU).

\subsection{Subtask C: Topic-Level Polarity}

\begin{table}[t!]
\begin{center}
\small{
\begin{tabular}{llcc}
 \hline
 \bf \# & \bf System & \bf Tweet & \bf Tweet\\
  & & & \bf sarcasm \\
 \hline
  1 & TwitterHawk & \bf 50.51$_1$ & 31.30$_2$ \\
  2 & KLUEless & \bf 45.48$_2$ & 39.26$_1$ \\
  3 & Whu-Nlp & \bf 40.70$_3$ & 23.37$_5$ \\
  4 & whu-iss & \bf 25.62$_4$ & 28.90$_4$ \\
  5 & ECNU & \bf 25.38$_5$ & 16.20$_6$ \\
  6 & WarwickDCS & \bf 22.79$_6$ & 13.57$_7$ \\
  7 & UMDuluth-CS8761 & \bf 18.99$_7$ & 29.91$_3$\\ \hline
& baseline & \bf 26.7 & 26.4\\
\hline
\end{tabular}
}
\caption{\textbf{Results for Subtask C: Topic-Level Polarity.}
             The systems are ordered by the official 2015 score.}
\label{tab:res-subtaskC}
\end{center}
\vspace*{-5mm}
\end{table}

The results for subtask C are shown in Table~\ref{tab:res-subtaskC}.
This proved to be a hard subtask, and only three of the seven teams that participated in it
managed to improve over a majority vote baseline.
These three teams,
TwitterHawk (using subtask B data to help with subtask C) with $F_1$=50.51,
KLUEless (which ignored the topics as if it was subtask B) with $F_1$=45.48,
and
Whu-Nlp with $F_1$=40.70,
achieved scores that outperform the rest by a sizable margin:
15-25 points absolute more than the fourth team.

Note that, despite the apparent similarity,
subtask C is much harder than subtask B:
the top-3 teams achieved an $F_1$ of 64-65 for subtask B vs. an $F_1$ of 41-51 for subtask C.
This cannot be blamed on the class distribution,
as the difference in performance of the majority class baseline is much smaller:
30.3 for B vs. 26.7 for C.

Finally, the last column in the table reports the results for the 75 sarcastic 2015 tweets. The winner here is KLUEless with an $F_1$ of 39.26,
followed by TwitterHawk with $F_1$=31.30,
and then by UMDuluth-CS8761 with $F_1$=29.91.

\subsection{Subtask D: Trend Towards a Topic}

\begin{table}[t!]
\begin{center}
\small{
\begin{tabular}{llcc}
\hline
\bf \# & \bf Team & \bf avgDiff	& \bf avgLevelDiff\\ 
\hline
   1 & KLUEless & \bf 0.202 & 0.810\\
   2 & Whu-Nlp & \bf 0.210 & 0.869\\
   3 & TwitterHawk & \bf 0.214 & 0.978\\
   4 & whu-iss & \bf 0.256 & 1.007\\
   5 & ECNU & \bf 0.300 & 1.190\\
   6 & UMDuluth-CS8761 & \bf 0.309 & 1.314\\
\hline
     & baseline & \bf 0.277 & 0.985\\
\hline
\end{tabular}
}
\caption{\textbf{Results for Subtask D: Trend Towards a Topic.}
             The systems are sorted by the official 2015 score.}
\label{tab:res-subtaskD}
\end{center}
\vspace*{-5mm}
\end{table}

The results for subtask D are shown in Table~\ref{tab:res-subtaskD}.
This subtask is closely related to subtask C
(in fact, one obvious way to solve D is to solve C and then to calculate the proportion),
and thus it has attracted the same teams, except for one.
Again, only three of the participating teams managed to improve over the baseline;
not suprisingly, those were the same three teams that were in top-3 for subtask C.
However, the ranking is different from that in subtask C,
e.g., TwitterHawk has dropped to third position,
while KLUEless and Why-Nlp have each climbed one position up
to positions 1 and 2, respectively.

Finally, note that avgDiff and avgLevelDiff yielded the same rankings.

\subsection{Subtask E: Degree of Prior Polarity}

Ten teams participated in subtask E. 
%Only one run was allowed for each team. 
Many chose an unsupervised approach and leveraged newly-created and pre-existing sentiment lexicons
such as 
the Hashtag Sentiment Lexicon,
the Sentiment140 Lexicon \cite{NRCJAIR14},
the MPQA Subjectivity Lexicon \cite{Wilson05},
and SentiWordNet \cite{swn}, among others. 
Several participants further automatically created their own sentiment lexicons from large collections of tweets.
Three teams, including the winner INESC-ID, adopted a supervised approach and used word embeddings (supplemented with lexicon features) to train a regression model.

The results are presented in Table~\ref{tab:res-subtaskE}.
The last row shows the performance of a lexicon-based baseline. 
For this baseline, we chose the two most frequently used existing, publicly available, and automatically generated sentiment lexicons: Hashtag Sentiment Lexicon and Sentiment140 Lexicon \cite{NRCJAIR14}.\footnote{\url{http://www.purl.com/net/lexicons}}
These lexicons have real-valued sentiment scores for most of the terms in the test set.
For negated phrases, we use the scores of the corresponding negated entries in the lexicons.
For each term, we take its score from the Sentiment140 Lexicon if present; otherwise, we take the term's score from the Hashtag Sentiment Lexicon.
For terms not found in any lexicon, we use the score of 0, which indicates a neutral term in these lexicons.
The top three teams were able to improve over the baseline.

\setlength{\tabcolsep}{4pt}

\begin{table}[t!]
\begin{center}
\small{
\begin{tabular}{lcc}
\hline {\bf Team}	& {\bf Kendall's $\tau$} & {\bf	Spearman's $\rho$}\\ 
& {\bf coefficient} & {\bf	coefficient}\\ \hline
INESC-ID &	\textbf{0.6251}	& 0.8172\\
lsislif	& \textbf{0.6211}	& 0.8202\\
ECNU	& \textbf{0.5907}	& 0.7861\\
CLaC-SentiPipe	& \textbf{0.5836}	& 0.7771\\
KLUEless	& \textbf{0.5771}	& 0.7662\\
UMDuluth-CS8761-10	& \textbf{0.5733}	& 0.7618\\
IHS-RD-Belarus	& \textbf{0.5143}	& 0.7121\\
sigma2320	& \textbf{0.5132}	& 0.7086\\
iitpsemeval	& \textbf{0.4131}	& 0.5859\\
RGUSentminers123	& \textbf{0.2537}	& 0.3728\\[4pt] \hline
Baseline & \textbf{0.5842} & 0.7843 \\
\hline
\end{tabular}
}
\caption{\textbf{Results for Subtask E: Degree of Prior Polarity.} The systems are ordered by their Kendall's $\tau$ score, which was the official score.}
\label{tab:res-subtaskE}
\end{center}
\vspace*{-5mm}
\end{table}

\section{Discussion}

As in the previous two years, almost all systems used supervised learning.
Popular machine learning approaches included SVM, maximum entropy, CRFs, and linear regression.
In several of the subtasks, the top system used deep neural networks and word embeddings,
and some systems benefited from special weighting of the positive and negative examples.

Once again, the most important features were those derived from sentiment lexicons.
Other important features included bag-of-words features, hashtags, handling of negation, word shape and punctuation features, elongated words, etc.
Moreover, tweet pre-processing and normalization were an important part of the processing pipeline.

Note that this year we did not make a distinction between constrained and unconstrained systems,
and participants were free to use any additional data, resources and tools they wished to.

Overall, the task has attracted a total of 41 teams,
which is comparable to previous editions:
there were 46 teams in 2014, and 44 in 2013.
As in previous years, subtask B was most popular,
attracting almost all teams (40 out of 41).
However, subtask A attracted just a quarter of the participants (11 out of 41),
compared to about half in previous years,
most likely due to the introduction of two new, very related subtasks C and D (with 6 and 7 participants, respectively).
There was also a fifth subtask (E, with 10 participants),
which further contributed to the participant split.

We should further note that our task was part of a larger Sentiment Track,
together with three other closely-related tasks,
which were also interested in sentiment analysis:
Task 9 on CLIPEval Implicit Polarity of Events,
Task 11 on Sentiment Analysis of Figurative Language in Twitter,
and Task 12 on Aspect Based Sentiment Analysis.
Another related task was Task 1 on Paraphrase and Semantic Similarity in Twitter,
from the Text Similarity and Question Answering track,
which also focused on tweets.

\section{Conclusion}

We have described the five subtasks organized as part of SemEval-2015 Task 10 on Sentiment Analysis in Twitter: detecting sentiment of terms in context (subtask A), classifiying the sentiment of an entire tweet, SMS message or blog post (subtask B), predicting polarity towards a topic (subtask C), quantifying polarity towards a topic (subtask D), and proposing real-valued prior sentiment scores for Twitter terms (subtask E). Over 40 teams participated in these subtasks, using various techniques.
%and resources.
%, which we also summarized.

We plan a new edition of the task as part of SemEval-2016,
where we will focus on sentiment with respect to a topic,
but this time on a five-point scale,
which is used for human review ratings on popular websites
such as Amazon, TripAdvisor, Yelp, etc.
From a research perspective, moving to an ordered five-point scale
means moving from binary classification to \emph{ordinal regression}.

We further plan to continue the trend detection subtask,
which represents a move from classification to \emph{quantification},
and is on par with what applications need.
They are not interested in the sentiment of a particular tweet
but rather in the percentage of tweets that are positive/negative.

Finally, we plan a new subtask on trend detection, but using a five-point scale,
which would get us even closer to what business (e.g. marketing studies),
and researchers, (e.g. in political science or public policy), want nowadays.
From a research perspective, this is a problem of \emph{ordinal quantification}.

\section*{Acknowledgements}

The authors would like to thank SIGLEX for supporting subtasks A--D,
and the National Research Council Canada for funding subtask E.

%\textbf{1. Do we need to acknowledge funding for Sarcasm this year?}

%\textbf{2. We need to fix the references -- some lack page numbers, etc.}

\bibliographystyle{naaclhlt2015}
\bibliography{naacl2015}

\end{document}